\documentclass{article}

\usepackage[final]{corl_2020} % Uncomment for the camera-ready ``final'' version.

\usepackage{amsmath, amssymb}
\usepackage{comment}

\usepackage{cleveref}
\crefname{figure}{Fig.}{Fig.}
\Crefname{figure}{Figure}{Figures}
\crefname{equation}{}{}
\Crefname{equation}{Equation}{Equations}

\usepackage{tikz}
\usetikzlibrary{positioning, shapes.geometric}
\usepackage[font=small,labelfont=bf]{caption}
\usepackage{subcaption}
\usepackage{algorithm}
\usepackage{algorithmic}
\usepackage{pgfplots}
\usepackage{enumitem}

\usepackage{mathtools}
\usepackage{tikz}
\newcommand{\Ngon}[2][]{\vcenter{\hbox to 0.1cm{\begin{tikzpicture}
\node[regular polygon,regular polygon sides=#2,draw,minimum size=0.001cm,#1](#2-gon){};
\end{tikzpicture}}}}

\title{Tolerance-Guided Policy Learning for Adaptable and Transferrable Delicate Industrial Insertion}

\author{
  Boshen Niu\thanks{These authors contributed equally to this article. The order of authorship was determined by a coin flip.}, Chenxi Wang\footnotemark[1], Changliu Liu\\
  Carnegie Mellon University, USA\\
  \texttt{bniu, chenxiwa, cliu6@andrew.cmu.edu}
}

\begin{document}
\maketitle

%===============================================================================

\begin{abstract}
    Policy learning for delicate industrial insertion tasks (e.g., PC board assembly) is challenging. This paper considers two major problems: how to learn a diversified policy (instead of just one average policy) that can efficiently handle different workpieces with minimum amount of training data, and how to handle defects of workpieces during insertion. To address the problems, we propose tolerance-guided policy learning. 
    To encourage transferability of the learned policy to different workpieces, we add a task embedding to the policy's input space using the insertion tolerance. Then we train the policy using generative adversarial imitation learning with reward shaping (RS-GAIL) on a variety of representative situations. 
    To encourage adaptability of the learned policy to handle defects, we build a probabilistic inference model that can output the best inserting pose based on failed insertions using the tolerance model. The best inserting pose is then used as a reference to the learned policy. 
    This proposed method is validated on a sequence of IC socket insertion tasks in simulation. The results show that 1) RS-GAIL can efficiently learn optimal policies under sparse rewards; 2) the tolerance embedding can enhance the transferability of the learned policy; 3) the probabilistic inference makes the policy robust to defects on the workpieces.
\end{abstract}


%===============================================================================

\section{Introduction}
    %============= Problem multi-pin-hole insertion
    Although robotics have been widely applied in manufacturing, it remains challenging for robots to handle delicate industrial insertion \cite{yun2008compliant,chhatpar2001search,jain2013scara}. This paper mainly considers insertion tasks on a small scale with multiple contact points. For example, the assembly of an IC socket that has multiple pins on a printed circuit board, the plug of a USB port, etc. 
    %"or different shapes", since polygons do not have multiple pins
    %============= Existing solutions
    Conventional model-based methods for insertion are able to achieve high performance on modeled tasks, but require case-by-case tuning and have poor generalizability \cite{qiao1993robotic}. As the emphasis of electronic assembly shifts from massive production to massive customization, the assembly lines are made more and more flexible, which requires robots to be equipped with general insertion skills that cover a wide variety of tasks. Learning-based methods are promising for robots to learn those skills \cite{fan2019learning,inoue2017deep,luo2019reinforcement,schoettler2020meta}.
    %Traditionally people use predefined trajectory or position controller for assembly tasks, which is sensible to noise and cannot be generalized to different workpieces. 
    %Learning based methods are promising for automatically getting a control strategy. \cite{Fan2018}\cite{Inoue2017}\cite{Luo2019}\cite{Schoettler2020}\cite{Yun2008}  
    %============= Challenges: diversity of workpieces, defects
    Nonetheless, there are two major challenges for policy learning in delicate industrial insertion tasks. The first is how to learn a diversified policy (instead of just one average policy) that can efficiently handle different workpieces with minimum amount of training data. 
    The second is how to handle defects of workpieces during insertion, e.g., bent pins.
    
    %============= Proposed solution: tolerance-guided policy learning
    This paper proposes a tolerance-guided policy learning method to address the problems as a way to leverage models (i.e., tolerance) in the policy learning. Tolerance refers to the amount of misalignment error and force error that is admissible so that the insertion can still succeed. Every insertion task has a tolerance. Different insertion tasks may have the same amount of tolerance. We can explicitly explore the similarities in the tolerance to transfer learned policies to different insertion tasks. The (nominal) tolerance can be directly computed from the CAD models and the specs of the insertion tasks. We call the mathematical description of the tolerance as a \textit{tolerance model}. To encourage transferability of the policy, we augment the policy input space to include the tolerance model and train the policy on a variety of representative tasks. To remove redundant information in the tolerance model, we encode the tolerance model into a low dimensional vector using auto-encoder. With the tolerance embedding, the next problem is how to efficiently learn the policy as the insertion tasks are with sparse rewards. By leveraging imitation learning and reinforcement learning, we propose generative adversarial imitation learning with reward shaping (RS-GAIL) that starts the learning by mimicking a suboptimal expert and gradually converges to a policy that is optimal with respect to the environment reward. The expert effectively guides the initial exploration. In addition, to encourage adaptability of the learned policy to handle defects, we build a probabilistic inference model that can output the best inserting pose based on failed insertions. The probabilistic inference starts from the nominal tolerance model (computed from CADs) and will gradually learn the true tolerance model (with defects). The best inserting pose based on the true tolerance model is then used as a reference to the learned policy. 
    %============= Contribution: 1,2,3
    The contributions of the paper include:
    \begin{itemize}[leftmargin=*,noitemsep]
        \item introduction of RS-GAIL to efficiently train insertion policies in sparse reward environments;
        \item policy generalization to different workpieces (with different number of pins and different geometry of the pins) using tolerance embedding;
        \item introduction of probabilistic inference of the defects on workpieces and the consequent optimal insertion points to make the policy robust to defects.
    \end{itemize}
    
    %The remainder of the paper is organized as follows. \Cref{sec: related work} discusses related work in policy learning. \Cref{sec: problem} formulates the tolerance model and introduces the system architecture. \Cref{sec: learning} introduces the RS-GAIL algorithm for policy training. \Cref{sec: adaptation} discusses the adaptation of the policy using tolerance embedding and probabilistic inference. \Cref{sec: result} discusses the experiment setup and result. Limitations and possible extensions of the work are discussed in \Cref{sec: discussion}. \Cref{sec: conclusion} concludes the paper.

%===============================================================================

\section{Related Work\label{sec: related work}}
    %============= Algorithm wise
    % Policy learning is to learn a neural network with state input and action output. Imitation learning (IL) and reinforcement learning (RL) are widely used in policy training. Behavior cloning (BC) \cite{Bojarski2016} trains the policy network using expert states and actions collected in the environment by supervised learning. However, behavior cloning requires large amount of expert data, and the performance of the learned policy is not likely to be better than the expert policy. Dataset Aggregation (DAGGER) \cite{Ross2011} mitigates BC's shortage of expert data by collecting data using the currently policy, but it has to get access the expert to label the corresponding actions of the generated states. Deep deterministic policy gradient (DDPG) \cite{Lillicrap2015} is an actor-critic based, model-free RL algorithm, and is the state-of-the-art RL algorithm for solving tasks with continuous action space. However, if the reward is sparse, such as the assembly tasks with strict tolerance, expensive computation is required for DDPG to explore, obtain the sparse reward, and converge.
\paragraph{Learning Framework for Insertion}
    An insertion policy takes the state (e.g., pin/hole relative pose, contact force) as input, and outputs an action for the robot to move the workpiece. 
    Inoue et al \cite{inoue2017deep} used LSTM based policy networks to map the current contact force and peg position to desired force and peg position. Luo et al \cite{luo2019reinforcement} used model-based RL algorithm and iLQG to train a torque controller and used a neural network to process force/torque readings as reference inputs to the controller. These works focus on single task without workpiece variations, hence the learned policy cannot generalize to new workpieces. Using meta-reinforcement learning \cite{schoettler2020meta,rakelly2019efficient}, the robot can learn skills on new tasks with a small amount of demonstrations. However, since the specifications of workpieces (e.g., number of pins, shape of pins) have large variations, it is intractable to get demonstrations for all new workpieces.
    Our proposed method, by encoding task specifications into low dimensional parameters and embedding these parameters as another input modality of the policy network, can generalize to new workpieces without training or finetuning using new data.
\paragraph{Learning Algorithm}
    To learn a policy, one can use either imitation learning (IL) or reinforcement learning (RL).
    In IL, the agent learns the policy by mimicking expert's behavior. Among common IL algorithms, behavior cloning (BC) \cite{bojarski2016end} requires large amount of expert demonstration data; dataset aggregation (DAGGER) \cite{ross2011reduction} requires access to the expert during rollout. Generative adversarial imitation learning (GAIL) \cite{ho2016generative} can overcome these problems. It includes a discriminator to measure similarity between the student policy and the expert policy, and uses the similarity as criteria to improve the student policy (similar to a RL problem). However, since IL training drives the state-action distribution of student policy towards that of the expert, it is impossible for the student to outperform the expert. In RL, the agent learns the policy by interacting with the environment according to a reward function which needs to be carefully engineered. For tasks with sparse rewards (such as the insertion task considered in this paper), RL algorithms (e.g., DDPG \cite{lillicrap2015continuous}) converge slowly and are not data-efficient due to the difficulty to explore non-zero rewards. To address the problem, we may add demonstration data into the replay buffer \cite{vecerik2017leveraging,nair2018overcoming,kim2013learning,hester2017deep}. However, these algorithms either need a large amount of demonstration data to balance the data distribution, or may still diverge due to the difficulty in exploration. 
    On the other hand, adding priors to current reward function, namely reward shaping (RS) \cite{syed2012imitation,judah2014imitation,bhattacharyya2019simulating, 2020arXiv201101298W} can guide the policy towards the desired behaviors indicated by the priors. By adding RS to IL, the student policy can explore more efficiently due to the guidance provided by the expert, and keep the potential to capture the desired behaviors to outperform the expert. Our proposed policy learning method leverages IL, RL, and RS, which is data-efficient, converges fast, and can outperform the suboptimal expert demonstrator in sparse reward environments.

\section{Problem Formulation\label{sec: problem}}
    %============= Insertion and tolerance
    %We consider the problem of training a diversified policy that can generate efficient adapted strategies for insertion tasks with different workpieces, transfer the learnt strategies to unseen workpieces with minimum amount of training data, and adapt to insertion tasks with defected workpieces. We describe  problem as:

    This paper considers the delicate industrial insertion tasks. 
    Let $\varepsilon$ be a task (characterized by the workpiece) and $\mathcal{E}$ be the distribution of tasks. The goal is to efficiently learn a policy $\pi_\varepsilon$, which varies according to the tasks, that achieves the highest reward over the distribution of tasks. 
    Let $\tau$ be the trajectory generated by $\pi_\varepsilon$ and $R(\tau)$ be the reward function on $\tau$, which evaluates the completion of the task within a given time frame, the duration of the task, and the number of collisions during the task. Then the policy learning problem can be written as the following optimization:
    \begin{equation}
        \max_{\pi_\varepsilon} ~~ \mathbb{E}_{\varepsilon \sim \mathcal{E}} [~\mathbb{E}_{\tau \sim \pi_\varepsilon}[R(\tau) ] ~].
        \label{eqn:main problem}
    \end{equation}
    It is worth noting that every task has a nominal model that is characterized by the design parameters of the workpiece (e.g., CAD model), while the real workpiece may deviates from the design parameters due to defects, which are difficult to be perceived before insertion. We define a mapping from a real workpiece $\varepsilon$ to its nominal model $\bar\varepsilon$ as $f:\varepsilon \mapsto \bar\varepsilon$. It is assumed that before insertion, only the nominal model $\bar\varepsilon$ is available, while we can infer the actual model $\varepsilon$ based on the insertion performance. In the following discussion, we introduce the task embedding using tolerance.
    
\paragraph{Tolerance}
    The performance of policy depends on the system dynamics, which then depends on the task. It is natural to use the peculiar dynamic properties of different workpieces to encode various insertion tasks. Here we use tolerance as the fundamental property to construct task encoding. Tolerance $\mathcal{T}(\varepsilon)$ of task $\varepsilon$ is defined as the set of states $\bar{s}$ of the workpiece so that the pins (denoted as the set $\mathcal{C}_{p,\varepsilon}(\bar{s})$) can be inserted into the holes (denoted as the set $\mathcal{C}_{h,\varepsilon}$):
    \begin{equation}
        \mathcal{T}(\varepsilon) = \{~ \bar{s} ~ | ~ \mathcal{C}_{p,\varepsilon}(\bar{s}) \subset \mathcal{C}_{h,\varepsilon} ~\}.
        \label{eqn:tol defination}
    \end{equation}
    \Cref{fig:tol gt1,fig:tol gt2} show examples for tolerance. Workpieces with different specifications may result in the same tolerance. Tasks with similar tolerances should have similar policies. %It is comprehensible since normally the outer pin-hole pairs are the active constraints for insertion. Workpieces with the same tolerance can also share the same insertion strategy. Only the upper parts of tolerance are shown here since it can be proved that the difference of upper part and the lower part is negligible given small $\theta$. Meanwhile, the upper part is already distinctive enough to distinguish different workpieces.
    %We consider insertion as $\mathcal{C}_{p,\varepsilon} \subset \mathcal{C}_{h,\varepsilon}$ where $\mathcal{C}_{p,\varepsilon}$ is the set representing pins of the workpieces $\varepsilon$ and $\mathcal{C}_{h,\varepsilon}$ is the set representing corresponding holes, e.g. a vacant spatial cylinder for a hole. $\bar{s}$ denotes the state relating to insertions, e.g. relative poses of pins w.r.t. holes. Tolerance is defined as the set containing all states $\bar{s}$ with which pins can be inserted into holes. 
    However, since $\mathcal{T}(\varepsilon)$ is excessively redundant as task encoding, we will introduce an encoder to transfer tolerances to low-dimensional representations $\psi(\varepsilon)$ to be discussed in \cref{sec: embedding}. We then use $\psi(\varepsilon)$ to guide the policy in order to generate diversified strategies for different workpieces. Since the tolerance of defective realistic workpieces is inaccessible before insertion, we can only parameterize the policy using its nominal model $\bar{\varepsilon}=f(\varepsilon)$. Therefore, the policy learning problem in \cref{eqn:main problem} is decomposed into two problems: policy learning with respect to the nominal model formulated in \cref{eqn:main problem 2} and inference of the actual task $\varepsilon$.% (to be discussed in \cref{sec: learning} and \cref{sec: embedding}) 
    \begin{equation}
        \max_{\pi(\psi(\bar{\varepsilon}))} ~~ \mathbb{E}_{\varepsilon \sim \mathcal{E}} [~\mathbb{E}_{\tau \sim \pi(\psi(\bar{\varepsilon}))}[R(\tau) ] ~],
        \label{eqn:main problem 2}
    \end{equation}
    % (to be discussed in \cref{sec:probabilistic}).
    
\begin{figure}[t]
        \centering
        \begin{subfigure}{.3\textwidth}
            \begin{center}
                \includegraphics[width=5 cm, trim={5cm -2cm 1cm -2cm}, clip]{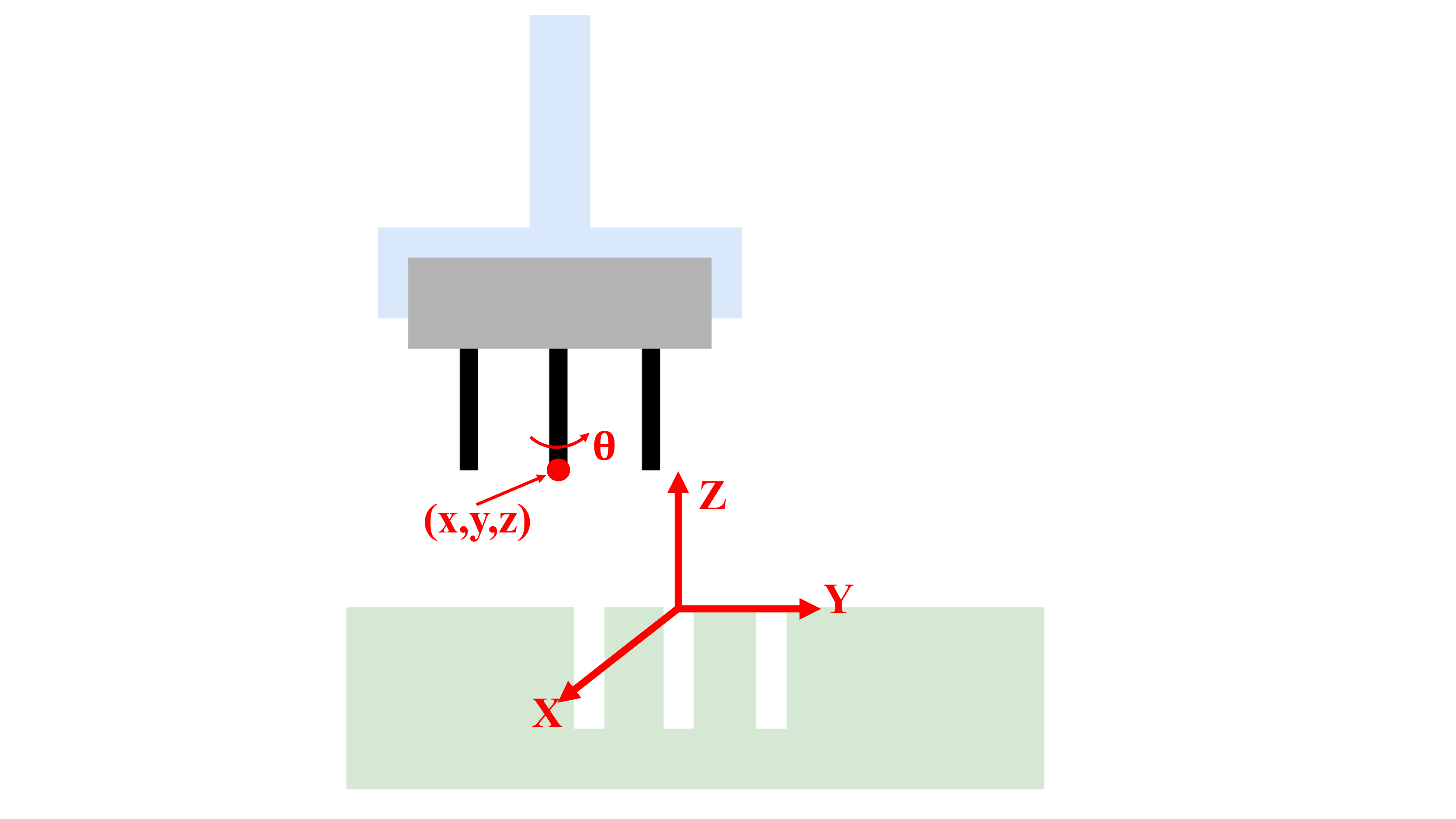}
            \end{center}
            \caption{}
            \label{fig: coord}
        \end{subfigure}
        \begin{subfigure}{.69\textwidth}
            \begin{center}
                \input{tikz/sys_diag_02}
                \caption{}
                \label{fig:diagram}
            \end{center}
        \end{subfigure}
        \begin{subfigure}{.99\textwidth}
            \begin{center}
                \begin{tikzpicture}
    % Design styles
    \def\x{0.75}
    \def\fz{\scriptsize}
    \definecolor{bl}{rgb}{0.109, 0.435, 0.890}
    \definecolor{gn}{rgb}{0.145, 0.894, 0.439}
    \definecolor{rd}{rgb}{0.917, 0.325, 0.364}
    \definecolor{cy}{rgb}{0.325, 0.917, 0.905}
    
    \tikzstyle{cl01} = [color=bl,fill=bl]
    \tikzstyle{cl02} = [color=gn,fill=gn]
    \tikzstyle{cl03} = [color=rd,fill=rd]
    \tikzstyle{cl04} = [color=cy,fill=cy]
    
    \tikzstyle{sz01} = [minimum height=0.8*\x cm , minimum width=1*\x cm]

    \tikzstyle{input} = [coordinate]
    
    % Define blocks
    \def\dist{4}
    \node[cl01, sz01, rounded corners, text=white] at (0,0) (pi_e) {\fz $\pi_e$};
    \node[cl01, sz01, rounded corners, text=white] at (0.8*\dist*\x,0) (pi_n_1) {\fz $\pi_{bc}$};
    \node[cl01, sz01, rounded corners, text=white] at (1.8*\dist*\x,0) (pi_n_2) {\fz $\pi_n$};
    \node[cl02, sz01, rounded corners, text=white] at (3*\dist*\x,0) (pi_phi_1) {\fz $\pi_{\phi}$};
    \node[cl03, sz01, rounded corners, text=white] at (4.1*\dist*\x,0) (pi_phi_2) {\fz $\pi_{\phi}^*$};
    
    % \node[cl04, sz01, rounded corners, text=white] at (0.4*\dist*\x,1*\x) (bc) {\tiny BC $|$ $\phi_1$ $|$ $\varepsilon_0$};
    
    % \node[cl04, sz01, rounded corners, text=white] at (1.3*\dist*\x,1*\x) (gail) {\tiny RS-GAIL $|$ $\phi_1$ $|$ $\varepsilon_0$};
    
    % \node[cl04, sz01, rounded corners, text=white] at (2.4*\dist*\x,1*\x) (te) {\tiny RS-GAIL $|$ $\phi_2$ $|$ $\varepsilon \sim \mathcal{E}_1$};
    
    % \node[cl04, sz01, rounded corners, text=white] at (3.5*\dist*\x,1*\x) (te) {\tiny Synthesis with PI};
     
     \draw[->, line width=0.5 mm] (pi_e) -- node[above] {\tiny BC $|$ $\phi_1$ $|$ $\varepsilon_0$} (pi_n_1);
     \draw[->, line width=0.5 mm] (pi_n_1) -- node[above] {\tiny RS-GAIL $|$ $\phi_1$ $|$ $\varepsilon_0$} (pi_n_2);
     \draw[->, line width=0.5 mm] (pi_n_2) -- node[above] {\tiny RS-GAIL $|$ $\phi_2$ $|$ $\varepsilon \sim \mathcal{E}_1$} (pi_phi_1);
     \draw[->, line width=0.5 mm] (pi_phi_1) -- node[above] {\tiny Synthesis with} node[below] {\tiny Probabilistic Inference} (pi_phi_2);
     
\end{tikzpicture}
                \caption{}
                \label{fig:train diagram}
            \end{center}
        \end{subfigure}
        \caption{\small System diagram. (a) Coordination system. (b) System architecture for tolerance-guided insertion. (c) Training pipeline for tolerance-guided policy. }
        \label{fig: overview}
        \vspace{-10pt}
    \end{figure}

    \paragraph{Assumptions and Notations}
    This paper focuses on the insertion tasks of circle-shaped multiple pin-hole pairs and polygon-shaped single pin-hole pairs. We use $n$*$m$ to denote that there are $n$ rows and $m$ pins per row. For example, \cref{fig: coord} shows a 1*3 case. The $x$ and $y$ axes span the horizontal plane and the $z$ axis is the vertical axis that aligns with the insertion direction. The origin is attached to the center of the board on its upper surface. It is assumed that pins and holes are rigid during insertion, and the rotations w.r.t. the $x$ or $y$ axis are negligible given that the robot controller is robust along these two axes. We denote $[x,y,\theta,z]$ as the pose of the workpiece, $[F_{x}, F_{y}, F_{z}]$ as the forces on the workpiece along each axis, $q_{\theta}$ as the torque on the workpiece along $\theta$, $s=[x,y,\theta,z, F_{x}, F_{y}, q_{\theta}, F_{z} ]$ as the state of workpiece, and $ a = [u_{x}, u_{y}, u_{\theta}, u_{z}] $ as velocities along each axis and also the action of the robot. $\bar{s}=[x,y,\theta]$ denotes the truncated state used to determine the tolerance and the success of insertions, i.e., if $\bar s\in\mathcal{T}(\varepsilon)$ and $z<0$, then the pins are inside the holes. The target insertion pose is denoted as $\bar s^* = [x^*,y^*,\theta^*]\in \mathcal{T}(\varepsilon)$. The coordinate system is defined such that the target insertion poses are $[0,0,0]$ for all nominal tasks. For real tasks, the target insertion pose may be set to a non-zero value to maximize the chance of successful insertion with defective workpieces. The measurement $\hat s = [\hat x,\hat y,\hat \theta,\hat z, \hat F_{x}, \hat F_{y}, \hat q_{\theta}, \hat F_{z}]$ of the state $s$ follows a Gaussian distribution with mean $s$. Define $\hat s^* :=[\hat x - x^*, \hat y - y^*, \hat\theta-\theta^*, \hat z, \hat F_x, \hat F_y,\hat q_\theta, \hat F_z]$ as the shifted measurement, which will be used as the input to the policy. We also define $r(s,a)$ as the reward function for state-action pairs, while the cumulative $r(s,a)$ over a trajectory $\tau$ is $R(\tau)$. The discount factor is set to be $1$.

    %============= Objective (minimize insertion time, minimize collision, handle defects)
    
    %============= System architecture
    \paragraph{System Architecture}
    
    The proposed policy $\pi_\varepsilon$ consists of two components  as shown in \cref{fig:diagram}: the policy $\pi(\psi(\bar{\varepsilon}))$ (shown as the policy network) and the probabilistic inference on the actual task $\varepsilon$ (shown as the outer loop). The policy $\pi(\psi(\bar{\varepsilon}))$ also contains two parts: the nominal policy (shown in blue) and the tolerance-guided adaptation (shown in green). The nominal policy $\pi_{n}: \hat s^* \rightarrow a$ considers only the measurement feedback and does not consider the diversity of the workpieces. It is trained on a representative workpiece $\varepsilon_{0}$. The nominal policy learns the common features of insertion tasks. The tolerance-guided adaptation is designed to handle distinctions among workpieces, serving similarly as learning residual \cite{johannink2019residual, silver2018residual}. It takes the task encoding $\psi(\bar{\varepsilon})$ as input and provides modifications to the nominal policy. We call the policy combining the nominal policy and the tolerance-guided adaptation as a diversified policy and parameterize the policy by $\phi$, i.e., $\pi_{\phi}(\psi(\bar{\varepsilon})):\hat s^*\times \psi(\bar{\varepsilon})\mapsto a$. The parameter contains two parts $\phi = [\phi_1; \phi_2$] where $\phi_1$ corresponds to the nominal policy and $\phi_2$ corresponds to the tolerance-guided adaptation. The training of  $\pi_{\phi}$ is discussed in \cref{sec: learning} and \cref{sec: embedding}. Probabilistic inference then exploits the insertion history of the current workpiece and infers the actual task $\varepsilon$ and the optimal reference point $\bar s^*$. The optimal reference for $\bar s^*$ will then be compensated to the input vector of $\pi_{\phi}$. The new policy is denoted $\pi_{\phi}^*$, to be discussed in \cref{sec:probabilistic}.

%===============================================================================
%\newpage
\section{Learning\label{sec: learning}}

    %============= RS-GAIL
    The learning algorithm will be used to train the nominal policy $\pi_{n}$ as well as the diversified policy $\pi_{\phi}$. The learning algorithm essentially solves the following problem
    \begin{equation}\label{eq: learning rl}
        \max_{\pi} \mathbb{E}_{\tau \sim \pi, \varepsilon}[R(\tau)].
    \end{equation}
    Since the workpieces have small tolerance in our application, positive rewards (i.e., successful insertion) are sparse. As a result, the exploration of RL algorithm becomes expensive. Imitation learning algorithms provide efficient guidance for exploration, but the performance of the learned policy is constrained by the optimality of the expert policy. We propose an integrated imitation learning and reinforcement learning framework, generative adversarial imitation learning with reward shaping (RS-GAIL), which leverages the effective exploration of IL and optimality of RL. Denote the expert policy as $\pi_e$ which may be generated by simple PID control or human demonstration. The original GAIL \cite{ho2016generative} solves the following minimax problem
    \begin{equation}
         \min_{\pi} \max_{D} \mathbb{E}_{\pi_{e}}[D(s,a)] - \mathbb{E}_{\pi}[D(s,a)],
    \label{eqn: GAIL minimax}
    \end{equation}
    where $D$ is the discriminator that tries to differentiate the learned policy from the expert policy. $s$ and $a$ are the states and actions generated by the policy. However, since the original GAIL does not contain the environment reward, the learned policy may be suboptimal. We then combine the two objective \eqref{eq: learning rl} and \eqref{eqn: GAIL minimax} through a weighting factor $\alpha\in[0,1]$:
    \begin{equation}
         \min_{\pi} \max_{D} (1-\alpha)\left[\mathbb{E}_{\pi_{e}}[D(s,a)] - \mathbb{E}_{\pi}[D(s,a)]\right] - \alpha \mathbb{E}_{\pi}[r(s,a)].
    \label{eqn: RS-GAIL minimax}
    \end{equation}
    When $\alpha = 0$, the problem reduces to GAIL, where $\pi$ goes to $\pi_{e}$. When $\alpha = 1$, the problem is a pure reinforcement learning problem, where $\pi$ optimizes the environment reward. In practice, we can start with a small $\alpha$ so that the policy $\pi$ can mimic $\pi_e$ to avoid useless exploration in the beginning, then gradually increase $\alpha$ in order to achieve optimality. 
    The detailed execution of the RS-GAIL algorithm is summarized in \cref{alg:rs_gail}.

    \begin{algorithm}[t]
        \caption{The Generative Adversarial Imitation Learning with Reward Shaping (RS-GAIL).}
        \label{alg:rs_gail} \footnotesize
        \begin{algorithmic}[1] 
    %%%%%%%%%%%%%%%%%%%%%%%%%\mathcal
    
        \STATE \textbf{Input:} Expert trajectory $ \tau_{e} \sim \pi_{e} $, initial policy parameters $ w_{init} $, reward shaping factor $\alpha$ \\
        \STATE Train $ \pi_{w_{init}} $ on $ \tau_{e} $ using Behavior Cloning, get $ \pi_{w_{BC}} $ \\
        \STATE Initialize the policy parameters in GAIL $ w_{0} $ with $ w_{BC} $, initialize discriminator parameters $ d_{0} $
        \FOR{$i=0,1,\hdots,N-1 $}
            \STATE Rollout student trajectory $ \tau_{i} \sim \pi_{\theta_{i}} $, updated the discriminator parameters from $ d_{i} $ to $ d_{i+1} $ using gradient descent with cost function
            \begin{equation}
                \max_{d_{i+1}} \mathbb{E}_{\pi_{e}}[D_{d_{i+1}}(s,a)] - \mathbb{E}_{\pi_{w_{i}}}[D_{d_{i+1}}(s,a)].
            \end{equation}
            \STATE Update the policy parameter from $ w_{i} $ to $ w_{i+1} $ using CMA-ES optimization \cite{hansen2016cma} on 
            % $ \alpha r(s,a) + (1-\alpha) \log(D_{d_{i+1}}(s,a)) $\\
            \begin{equation}
                \max_{w_{i+1}} (1-\alpha) \mathbb{E}_{\pi_{w_{i+1}}}[D_{d_{i+1}}(s,a)] + \alpha \mathbb{E}_{\pi_{w_{i+1}}}[r(s,a)].
            \end{equation}
            
        \ENDFOR
        \RETURN optimal policy $ \pi_{\theta_{N}} $
        \end{algorithmic}
    \end{algorithm}
%===============================================================================
\section{Adaptation\label{sec: adaptation}}
    %============= Tolerance-guided policy
    \begin{figure}[t]
        \centering
            \begin{subfigure}{0.4\textwidth}
                \begin{center}
                    \input{tikz/encoder}
                \end{center}
                \caption{Auto-encoder architecture}
            \end{subfigure}
            \begin{subfigure}{0.2\textwidth}
                \centering
                \begin{subfigure}{1\textwidth}
                    \begin{center}
                        \includegraphics[height=1.5 cm, trim={3.2cm 1.7cm 1.1cm 2.5cm}, clip]{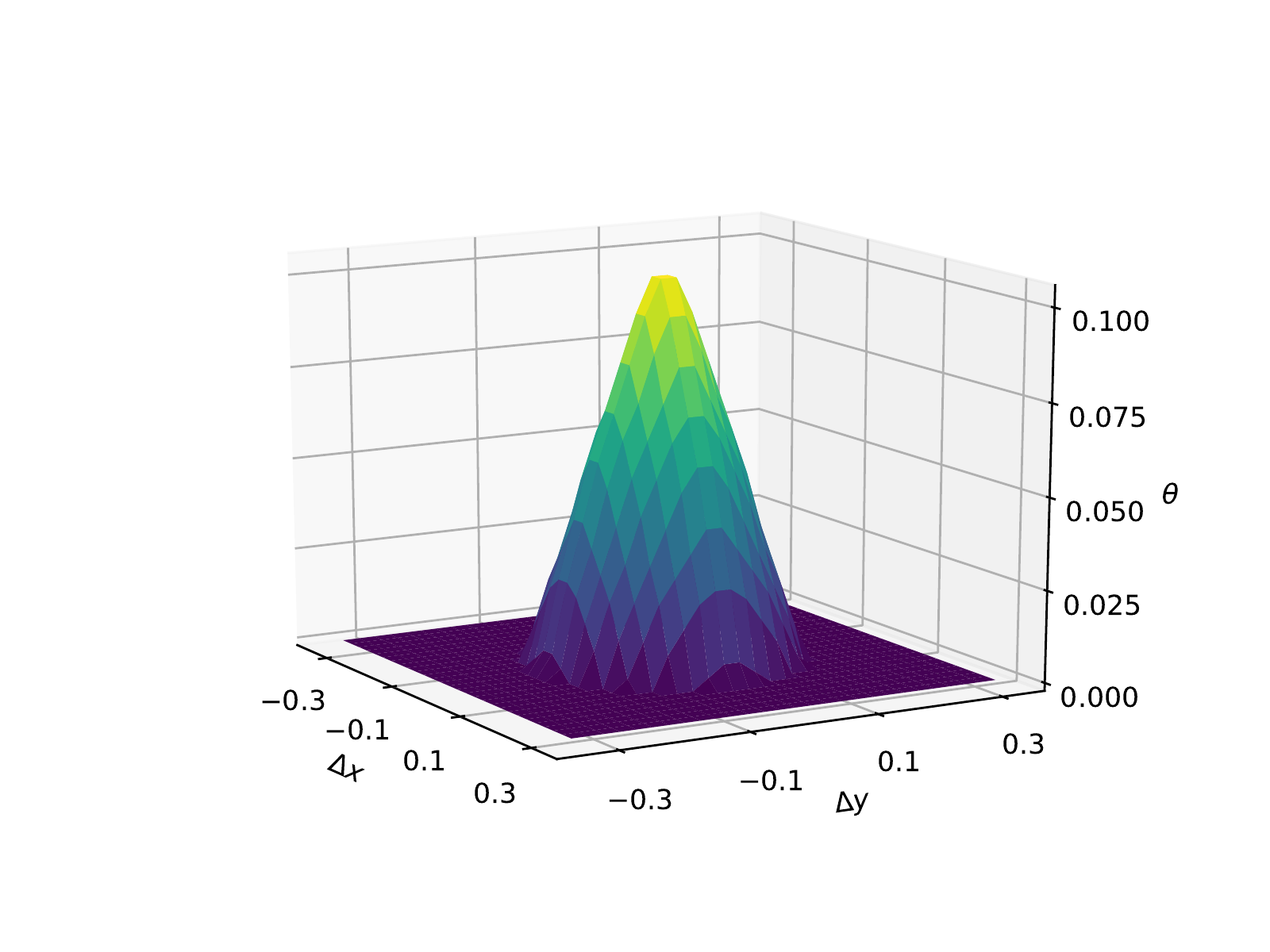}
                    \end{center}
                    \caption{Ground truth}
                    \label{fig:tol gt1}
                \end{subfigure}
                \begin{subfigure}{1\textwidth}
                    \begin{center}
                        \includegraphics[height=1.5 cm, trim={3.2cm 1.7cm 1.1cm 2.5cm}, clip]{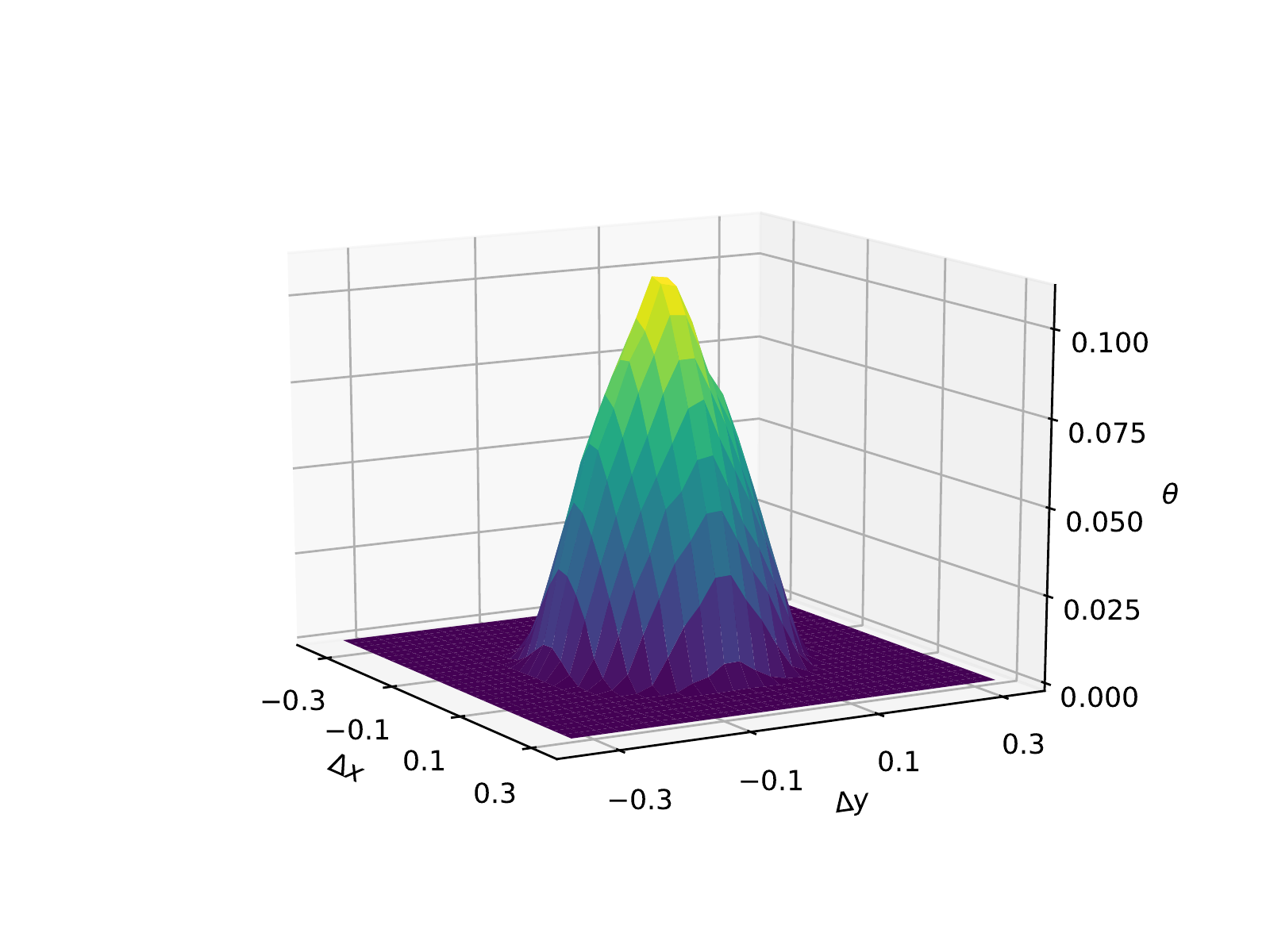}
                    \end{center}
                    \caption{Prediction}
                \end{subfigure}
            \end{subfigure}
            \begin{subfigure}{0.2\textwidth}
                \centering
                \begin{subfigure}{1\textwidth}
                    \begin{center}
                        \includegraphics[height=1.5 cm, trim={3.2cm 1.7cm 1.1cm 2.5cm}, clip]{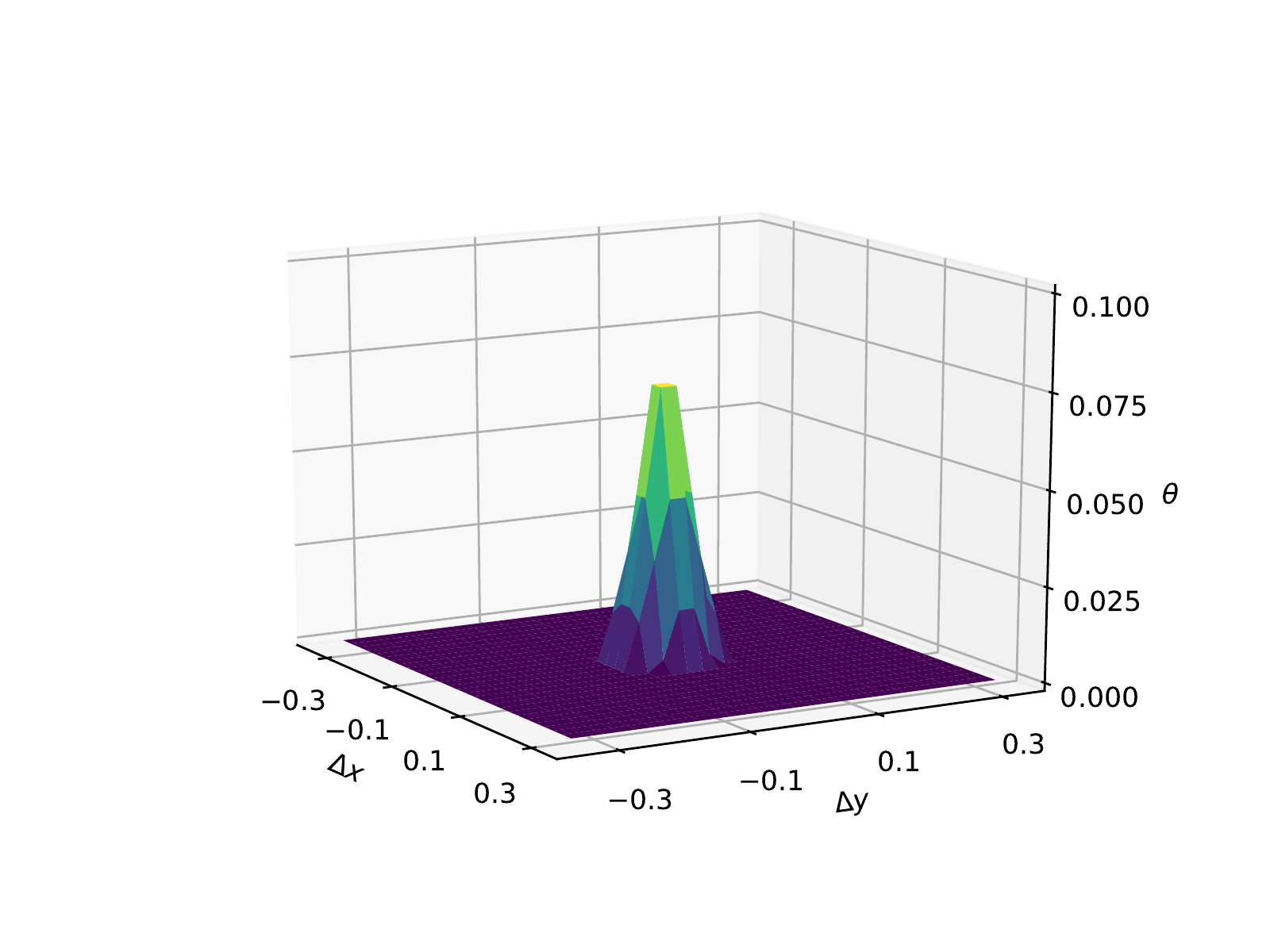}
                    \end{center}
                    \caption{Ground truth}
                    \label{fig:tol gt2}
                \end{subfigure}
                \begin{subfigure}{1\textwidth}
                    \begin{center}
                        \includegraphics[height=1.5 cm, trim={3.2cm 1.7cm 1.1cm 2.5cm}, clip]{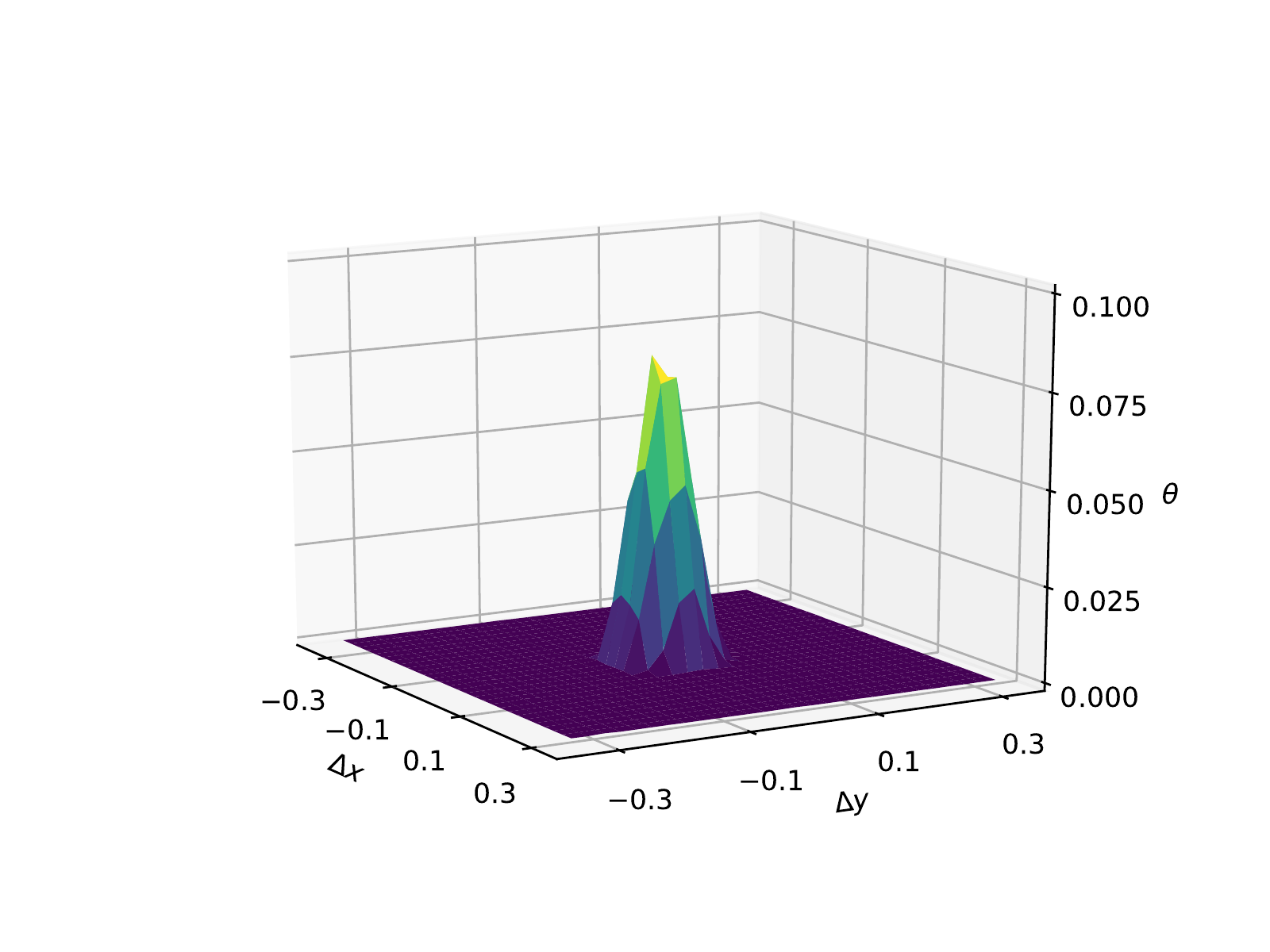}
                    \end{center}
                    \caption{Prediction}
                \end{subfigure}
            \end{subfigure}
        \caption{\small Tolerance embedding. (a) Architecture of the tolerance auto-encoder where cyan, green, and blue layers denote input, convolutional, and fully-connected layers. (b) Tolerance of a circle-shaped 2*2 workpiece with 0.3mm radius of pins, 0.5mm radius of holes, and 2.5mm as row/column intervals. (d) Tolerance of a regular-polygon-shaped 1*1 workpiece with 1.5mm circumradius of pins and 1.575mm circumradius of holes. (c) and (e) are the corresponding predictions. The units of $x$ and $y$ axes are 
       millimeter. The unit of $\theta$ axis is radius. Only the upper parts of tolerance are shown here since  the difference between the upper part ($\theta >0$) and the lower part ($\theta <0$) is negligible for small $\theta$. Meanwhile, the upper part is already distinctive enough to distinguish different workpieces. }
        \label{fig:auto-encoder} 
        \vspace{-10pt}
    \end{figure}
    
    \subsection{Tolerance-guided Policy}
    \label{sec: embedding}
    
    Convolutional auto-encoder \cite{lecun1995convolutional,hinton2006reducing} is introduced for task encoding $\psi(\varepsilon)$ to reduce the redundancy of tolerance $\mathcal{T}(\varepsilon)$ for $\varepsilon \sim \mathcal{E}$  as shown in  \cref{fig:auto-encoder}. Though we may directly encode the three-dimensional tolerance \cite{zhu2016deep,ji2017learning}, it is more efficient to do two-dimensional encoding by `flattening' the surface of the tolerance, i.e., using a value table to store $\theta$ on the tolerance surface and turning the volume of $\mathcal{T}(\varepsilon)$ to an image. In this way, we can use a solid simple-structured convolutional auto-encoders for two-dimensional image processing \cite{bengio2012unsupervised}. Encoders for circle-shaped workpieces and polygon-shaped workpieces are trained separately with corresponding training data. Training data for circle-shaped workpieces is sampled from 2*1 to 2*20 sockets with different radius of pins and radius of holes. Training data for polygon-shaped workpieces is sampled from 1*1 socket with shapes of triangle, rectangle, pentagon, and hexagon in different sizes. As shown in \cref{fig:auto-encoder}, we can successfully reduce the dimension of tolerance to $5$. 
    
    The overall training pipeline of the tolerance-guided policy is shown in \cref{fig:train diagram}. It can be viewed as a form of curriculum learning \cite{bengio2009curriculum}. We first start with the nominal part of the diversified policy $\pi(\psi(\varepsilon))$ on a representative task $\varepsilon_0$ to learn the general features of insertions. In this phase, only the parameters related to the measurements, i.e., $\phi_1$, are updated. According to \cref{alg:rs_gail}, we first initialize $\pi_{bc}:\hat s^*\mapsto a$ by cloning an expert policy $ \pi_e $. Then we obtain the nominal policy $\pi_n:\hat s^*\mapsto a$ by applying RS-GAIL on top of $\pi_{bc}$. In the second phase, we enable policy adaptation with tolerance embedding to learn the distinction among various tasks. The parameters in $\phi_1$ are copied from $\pi_{n}$ and do not change during the training. Parameters that are related to tolerance embedding, i.e., $\phi_2$, are updated from zero initiation using RS-GAIL. Tasks $\varepsilon$ in this phase are sampled from a training task set $\mathcal{E}_1$. The resulting policy is denoted $\pi_{\phi}:\hat s^*\times \psi(\varepsilon)\mapsto a$.
    
    %============= Probabilistic inference
    \subsection{Probabilistic Inference}
    \label{sec:probabilistic}
    
    Since the tolerance of realistic workpieces $\mathcal{T}(\varepsilon)$ may not align with its nominal form $\mathcal{T}(\bar{\varepsilon})$, the diversified policy $\pi(\psi(\bar{\varepsilon}))$ may deviate from the desired policy $\pi(\psi(\varepsilon))$. Therefore, we introduce probabilistic model to narrow the gap between the two policies. We leverage the information contained in previous insertions to infer the actual task $\varepsilon$ and the optimal insertion point $\bar s^*$. Denote $I_1,\hdots,I_m$ as the results of previous insertions and $\bar{s}_1,\hdots,\bar{s}_m$ as the corresponding insertion states. $m$ is the number of previous insertions. $I=1$ indicates successful insertion while $I=0$ indicates the opposite. Now we consider the problem of maximizing the expectation of $I_{m+1}$ over the next goal insertion state $\bar{s}^*$ given insertion histories: 
    $
        \max_{\bar s^*} ~ g_{m+1} = \mathbb{E} [I_{m+1} | I_1,\hdots,I_m, \bar{s}_1,\hdots,\bar{s}_m, \bar s^*].
        \label{eqn: obj pm}
    $
    We assume that the prior distribution of the realistic workpieces $\varepsilon$ given its nominal model $\bar{\varepsilon}$ is normal $\varepsilon \sim \mathcal{N}(\bar{\varepsilon}, \Sigma)$. The conditional probability $p( I_j ~|~ \bar{s}_j, \varepsilon )$ for the $j$th insertion attempt (denoted as $I_j$) equals to $I_j$ if $\mathcal{C}_{p,\varepsilon}(\bar{s}_j) \subset \mathcal{C}_{h,\varepsilon}$ or $1 - I_j$ if $\mathcal{C}_{p,\varepsilon}(\bar{s}_j) \not\subset \mathcal{C}_{h,\varepsilon}$
    Then we can infer the actual workpiece $\varepsilon$ using the previous insertion results and insertion states by applying the Bayesian rule:
    \begin{equation}
        p(\varepsilon ~|~ I_1,\hdots,I_m, \bar{s}_1,\hdots,\bar{s}_m)
        =
        \frac{\Pi_{j=1}^{m} p( I_j ~|~ \bar{s}_j, \varepsilon ) ~ p(\varepsilon)}
        {\int \Pi_{j=1}^{m} p( I_j ~|~ \bar{s}_j, \varepsilon ) ~ p(\varepsilon) ~ \mathrm{d}\varepsilon }.
        \label{eqn: inference}
    \end{equation}
    Then we can rewrite the objective $g_{m+1}$ in \cref{eqn: obj pm calculable}, which is conditioned on true insertion states. In practice, true states of workpieces are inaccessible. An altered objective $g'_{m+1}$ conditioned on measured states given the noise model $p(\bar{s} | \hat{\bar{s}})$ is shown in \cref{eqn: obj pm calculable noise}, where $\hat{\bar{s}}$ are the observed truncated states. 
    \begin{align}
        g_{m+1} &= \int p( I_{m+1} ~|~ \bar{s}^*, \varepsilon ) ~ p(\varepsilon ~|~ I_1,\hdots,I_m, \bar{s}_1,\hdots,\bar{s}_m) ~ \mathrm{d}\varepsilon
        \label{eqn: obj pm calculable},\\
        g_{m+1}^{'} &= \int g_{m+1} ~ p(\bar{s}_1,\hdots,\bar{s}_m ~|~ \hat{\bar{s}}_1,\hdots,\hat{\bar{s}}_m) ~ \mathrm{d}\bar{s}_1 \hdots \mathrm{d}\bar{s}_m
        \label{eqn: obj pm calculable noise}.
    \end{align}
    For real applications, both integrals in \cref{eqn: obj pm calculable} and \cref{eqn: obj pm calculable noise} are challenging to compute, which require complex slicing to transfer the multi-dimensional integral to practicable repeated integral. Therefore, we solve the problem numerically by sampling $\varepsilon$ from its distribution $N(\bar{\varepsilon}, \Sigma)$ and $\bar{s}$ from the measurement noise model. Various optimization methods can be applied to maximize $g_{m+1}$. Here we adopt the covariance matrix adaptation evolutionary method (CMA-ES) \cite{hansen2016cma}. At each iteration, we sample various $\bar{s}_{m+1}$ and evaluate them with the expectation in \cref{eqn: obj pm calculable} or \cref{eqn: obj pm calculable noise}. For a consecutive insertion task, we apply probabilistic inference whenever a collision occurs. With the knowledge of the insertion histories, the probabilistic inference utilizes CMA-ES to output the optimal goal insertion state $\bar{s}^*$. A workpiece can be considered as immoderately defected and better to be discarded if the optimal $g_{m+1}$ falls below a certain threshold, which can increase the robustness of industrial insertion production lines.

    An numerical study of the probabilistic inference is conducted on workpieces with 2*1 circle-shaped pins and random defects: stochastic horizontal translation of each pin. The workpiece has 0.3mm pin radius, 0.5mm hole radius, and 5mm nominal interval between two pins. This experiment only tests probabilistic inference and assumes perfect control to the goal position and perfect measurement of states. 500 samples are generated for the given distribution of defects where only around 60\% of samples can be inserted with unadapted goal states and 0.8\% of samples are impossible to insert regardless of goal states. We compare the probabilistic model with a random policy that randomly selects goals from a pre-defined set of goals. The results demonstrate that the probabilistic model outperforms random policy in terms of both the success rate within 10 attempts and the average insertion attempts before success. It increases the former from 60\% to 90.2\% and maintains an average attempts of 1.836. In contrast, the random policy has a success rate of 78.2\% and 2.796 average attempts.   

%===============================================================================

%\section{Experiment Set-up\label{sec: experiment}}

%===============================================================================
\section{Results and Discussions}
\label{sec: result}

    \paragraph{Experiment Setup}
    The proposed method has been tested in OpenAI Gym environment. The dynamics are specified in discrete time with sampling rate $dt = 0.25s$. When there is no collision, the movement follows $s' = s + a\cdot dt$. When there is a collision between two time steps, the robot will stop at the first collision point and receive a contact force (now set as a binary value) at the direction where the collision happens. In addition, the actions are saturated at $[-a_{\text{max}}, a_{\text{max}}]$.  
    %We derive the dynamics of the insertion process and implement it into an OpenAI Gym environment to test the performance of our algorithms. 
    %Given the actions $ a = [u_{x}, u_{y}, u_{\theta}, u_{z}]$ output by the policy, our environment first saturates the actions to the action space $[-a_{\text{max}}, a_{\text{max}}]$, where $ a_{\text{max}} $ to meet the actuation limit of real robot. The desired output state $ s^{\prime}_{\text{des}} $ os then calculated by the kinematics $ s^{\prime}_{\text{des}} = s + a \cdot dt_{\text{des}} $, where $dt_{\text{des}}$ is the step length. When collision happens, the workpiece stops at the collision position. In this case, a collison checker performs binary search between $ [0, dt_{\text{des}}] $ until find the collision boundary $ dt $, and calculates the output state by $ s^{\prime} = s + a \cdot dt $. The contact force is then updated by assign binary value ("1") to the direction where the collision happens. 
    At the start of each episode, the workpiece is reset to a random initial pose. %We add noise to the output state to simulate the observation noise.
    The reward function is defined to be $c(1-\frac{k}{k_{\text{max}}})$ if the insertion is finished before the maximum horizon $k_{\text{max}}$, and $-1$ if a collision happens. The parameter $c$ adjusts the trade-off between efficiency (fast insertion) and safety (collision minimization).
    %\begin{equation}
    %   r =     \begin{cases}       c(1-\frac{k}{k_{\text{max}}}) & \text{if at step $k$ ($k<k_{\text{max}}$), the insertion is finished}\\      
    %   -1 & \text{if collision happens in current step} \\
    %   0 & \text{otherwise}
    %   \end{cases},
    %\end{equation}
    %where $k_{\text{max}}$ is the maximum acceptable step to finish the insertion, $c$ is a weight parameter. 
    In our experiments, we set $ a_{\text{max}} = [2.5 \text{mm/s}, 2.5 \text{mm/s}, 0.1 \text{rad/s}, 2.5 \text{mm/s}] $, $ k_{\text{max}} = 40 $ and $c=100$. The standard deviation of the measurement noise is set to be $ [0.1 \text{mm}, 0.1 \text{mm}, 0.0 \text{rad}, 0.1 \text{mm}] $ for $ [x, y, \theta, z] $. For RS-GAIL training, the weight $\alpha$ in \eqref{eqn: RS-GAIL minimax} is set to $0.8$ and the maximum iteration is set to 100. The discriminator is updated once every 5 iterations. We stop training when the environment reward of the student policy stops increasing; and the probability that the student state-action pairs are classified by the discriminator as expert is stably around 40-60\%. 
    % We use $ dt=0.25 $, $c=100$, $k_{\text{max}}=40$ for our experiments. 
    % The performance evaluation metrics are defined as: accumulated reward, successful rate, number of steps to finish the insertion, and number of collisions. 

    We designed two expert policies, one emphasizing safety and the other emphasizing efficiency. The safe expert policy $\pi_e^{\text{safe}}$ first moves the workpiece horizontally to align the pins and holes, and then move vertically down to insert the workpiece. %This expert is optimal for collision avoidance, but suboptimal for efficiency and successful because it does not move straight from the initial pose to the desired pose.  
    The efficient expert policy $\pi_e^{\text{eff}}$ moves in full speed towards the hole. The safe expert has fewer collisions and longer insertion time than the efficient expert. The optimal behavior would be a balance between the two. %We then train the nominal policies $\pi_{bc}$ and $\pi_n$ from these two expert policies. 
    
\begin{figure}
        \centering
            \begin{subfigure}{0.49\textwidth}
                \begin{center}                             
                \includegraphics[width=8 cm, trim={4.5cm 2.7cm 1.1cm 2.5cm}, clip]{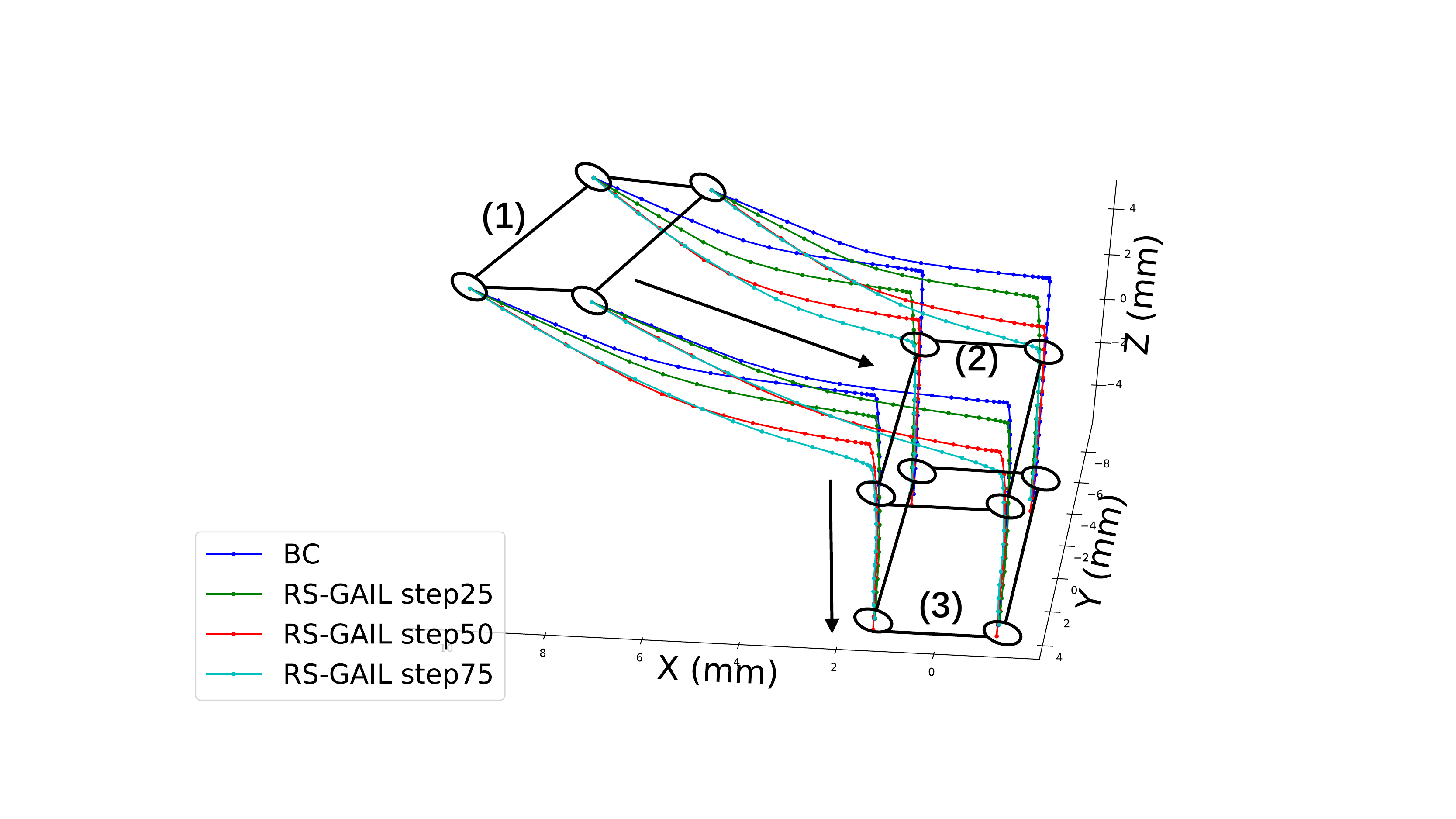}
                \end{center}
                \caption{}
                \label{fig:viz traj}
            \end{subfigure}
            \begin{subfigure}{0.49\textwidth}
                \begin{center}
                \includegraphics[width=6 cm, trim={2.5cm -1cm 3cm -1cm}, clip]{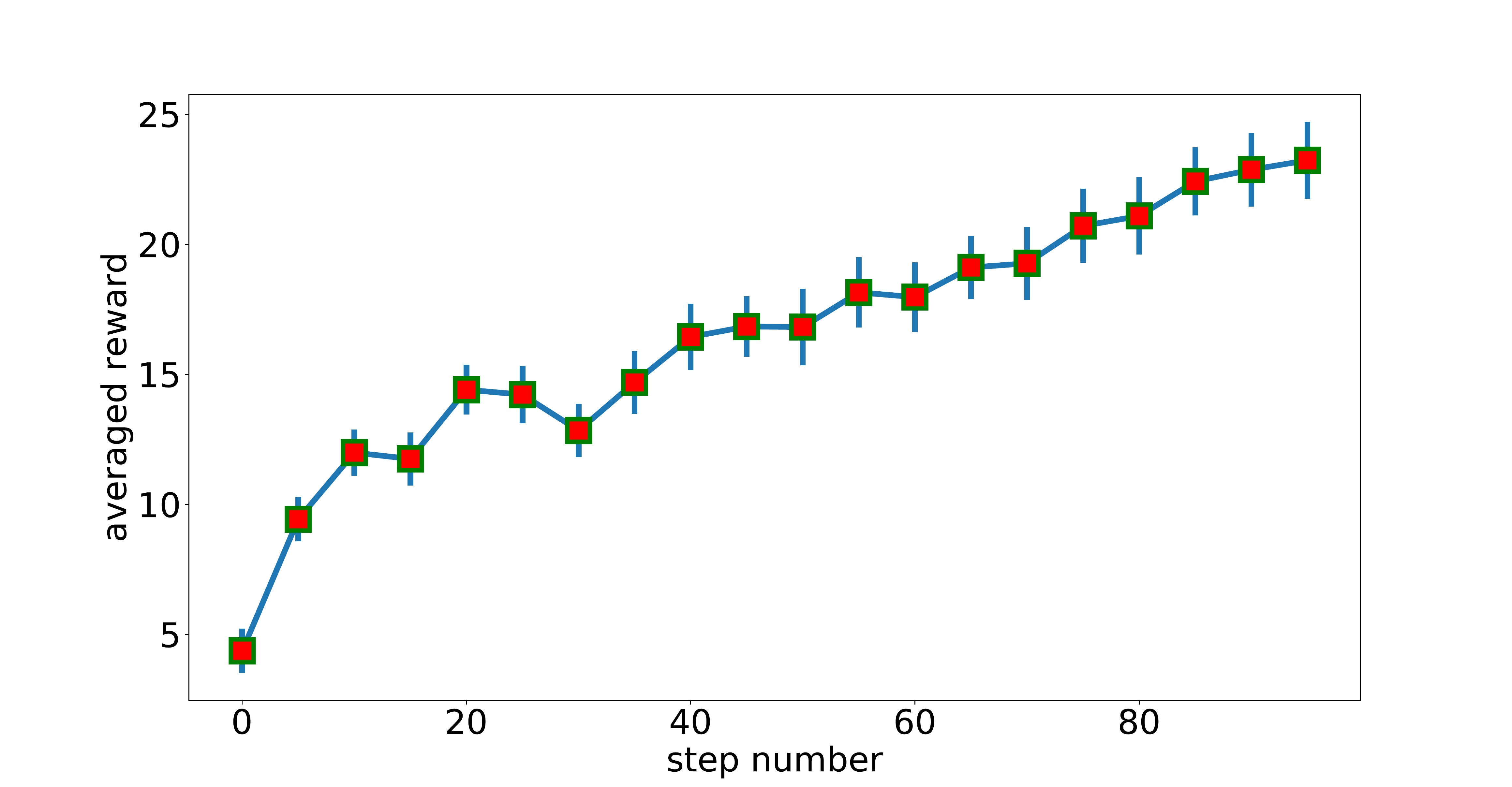}
                \end{center}
                \caption{}
            \end{subfigure}
        \caption{\small Illustration of the RS-GAIL training on the nominal policy in experiment C2.3. (a) Trajectory change over the training process demonstrated using a 2*2 workpiece, where (1) is the initial pin position; (2) is the pin position when the pins start entering the holes; and (3) is the pin position of successful insertion. (b) Reward (with variance) change over the training process in experiment C2.3.}
        \label{fig:rs-gail performance} 
    \end{figure}

    \paragraph{Results}
    \underline{Effect of different experts.} We trained nominal policies from the two experts $\pi_e^{\text{eff}}$ and $\pi_e^{\text{safe}}$. Comparing experiments C1.1 to C1.3 and C2.1 to C2.3 as shown in \cref{table:2*nCirclePerformance}, RS-GAIL optimizes the cloned policy $\pi_{bc}$ by reducing the number of collisions of $\pi_e^{\text{eff}}$ and speeding up the number of steps of $\pi_e^{\text{safe}}$. Because our RS-GAIL starts by mimicking the expert and then gradually optimizes the policy according to the environment rewards, even starting from experts with different behaviors, the policy can finally converge to similar optimal behavior. As shown in C1.3 and C2.3, we finally get similar $\pi_{n}$ with promising number of steps and number of collisions. 
    \underline{Effectiveness of RS-GAIL.} From \cref{fig:rs-gail performance} and C2.1 to C2.3, policies trained by behavior cloning $\pi_{bc}$ having similar behavior with the expert $\pi_e$, which first moves in xy plane for pin-hole alignment and then moves down for insertion. This `L' shaped trajectory is not optimal because longer travel distance induces more steps and less successful rate. As a result, $\pi_e$ gets low rewards, and $\pi_{bc}$ even lower. Our RS-GAIL starts from $\pi_{bc}$ and optimizes it according to our defined reward function. During the training process of RS-GAIL, the learned policy $\pi_{n}$ generates smoother trajectories, which have higher rewards and successful rate. We perform the following experiments using the $\pi_n$ in C2.3. 
    \underline{Generalization to new workpieces.} Consider C2.3, C3.1, and C3.2. When we introduce new workpiece 2*4 in C3.1, $\pi_{n}$ trained in C2.3 is not able to get good performance. However, our tolerance-guilded policy $\pi_{\phi}$ is able to handle this workpiece generalization problem. In C3.2, by training on data of workpiece 2*2 and 2*8 and only updating the parameters of the tolerance embedding networks ($\phi_{2}$), $\pi_{\phi}$ can be generalized to new unseen workpiece 2*4 and get higher performance than $\pi_{n}$. 
    \underline{Generalization to defective workpieces.} In C4.1 and C4.2, we utilize the same defective 2*1 workpieces in \cref{sec:probabilistic} to test synthesized probabilistic inference. Both the reward and the success rate drop significantly due to defects. The policy with probabilistic inference, i.e., $\pi_{\phi}^*$, mitigates the negative effects of defects and increases both the reward and the success rate. 
    \underline{Generalization to different pin geometries.} We also create a series of 1*1 workpieces with polygon-shape pins, including rectangle $\Ngon{4}$~~~, pentagon $\Ngon{5}$ ~~~, and hexagon $\Ngon{6}$~~~~. As shown in P1.1 to P1.3, RS-GAIL optimizes the policy similar to C2.1 to C2.3. Regardless of the differences in the shape of pin sections, those workpieces can be handled with similar policies. As a result, our robust nominal policy $\pi_{n}$ gets similar performance on $\Ngon{4}$~~~ and $\Ngon{5}$~~~~, as shown in P1.3 and P2.1. In P2.2, by training on $\Ngon{4}$~~~ and $\Ngon{6}$~~~, our $\pi_{\phi}$ slightly outperforms $\pi_{n}$ on $\Ngon{5}$~~~~.
    
    %============= Results
    % \begin{table}[ht]
    % \centering
    % \begin{tabular} {lllll}
    % Number & Network & Trainable parameters & Train env & Eval env \\\hline
    % 1 & $\pi_{n}$ ($\phi_{1}$) & $\phi_{1}$ & $\varepsilon$ & $\varepsilon$ \\
    % 2 & $\pi_{\Phi}$ ($\phi_{1}$, $\phi_{2}$) & $\phi_{2}$ & $\mathcal{E}_{1}$, $\mathcal{E}_{3}$ & $\mathcal{E}_{2}$ \\ \hline
    % \end{tabular}
    % \caption{\label{tab:widgets}Overview of experiment}
    % \end{table}
    \begin{table}
    \centering\scriptsize
    \begin{tabular} {llllllll}
    ID & Policy & Train env& Eval env & reward & successful rate & number of steps & number of collisions \\
    \hline
    C1.1 & $\pi_e^{\text{eff}}$ & 2*2 & 2*2 & $13.76\pm1.20$ & $0.75\pm0.03$ & $29.65\pm0.23$ & $6.11\pm0.71$ \\
    C1.2 & $\pi_{bc}$ & 2*2 & 2*2 & $6.57\pm3.07$ & $0.68\pm0.04$ & $30.91\pm0.71$ & $8.15\pm1.31$ \\
    C1.3 & $\pi_{n}$ & 2*2 & 2*2 & $\mathbf{20.82\pm0.76}$ & $\mathbf{0.94\pm0.03}$ & $\mathbf{31.05\pm0.24}$ & $\mathbf{1.69\pm0.28}$ \\ 
    \hline
    C2.1 & $\pi_e^{\text{safe}}$ & 2*2 & 2*2 & $11.28\pm0.72$ & $0.77\pm0.04$ & $35.49\pm0.29$ & $0.0\pm0.0$ \\
    C2.2 & $\pi_{bc}$ & 2*2 & 2*2 & $4.94\pm0.96$ & $0.58\pm0.05$ & $37.69\pm0.28$ & $0.83\pm0.31$ \\
    C2.3 & $\pi_{n}$ & 2*2 & 2*2 & $\mathbf{20.58\pm0.71}$ & $\mathbf{0.90\pm0.02}$ & $\mathbf{31.40\pm0.20}$ & $\mathbf{0.91\pm0.24}$ \\
    \hline
    C3.1 & $\pi_{n}$ & 2*2 & 2*4 & $16.38\pm1.53$ & $0.84\pm0.03$ & $33.37\pm0.43$ & $2.74\pm0.49$ \\
    C3.2 & $\pi_{\phi}$ & 2*2, 2*8 & 2*4 & $\mathbf{20.47\pm0.54}$ & $\mathbf{0.92\pm0.01}$ & $\mathbf{31.42\pm0.18}$ & $\mathbf{0.98\pm0.10}$ \\ 
    \hline
    C4.1 & $\pi_{\phi}$ & 2*2, 2*8 & 2*1 defected & $6.04\pm1.28$ & $0.30\pm0.03$ & $35.03\pm0.53$ & $6.42\pm0.11$ \\
    C4.2 & $\pi_{\phi}^*$ & - & 2*1 defected & $\mathbf{12.44\pm2.38}$ & $\mathbf{0.41\pm0.02}$ & $\mathbf{32.80\pm0.82}$ & $\mathbf{5.60\pm0.37}$ \\
    \hline
    P1.1 & $\pi_e^{\text{safe}}$ & $\Ngon{4}$ & $\Ngon{4}$ & $12.36\pm0.41$ & $0.82\pm0.03$ & $35.05\pm0.16$ & $0.0\pm0.0$ \\
    P1.2 & $\pi_{bc}$ & $\Ngon{4}$ & $\Ngon{4}$ & $7.46\pm0.30$ & $0.68\pm0.02$ & $37.02\pm0.12$ & $0.0\pm0.0$ \\
    P1.3 & $\pi_{n}$ & $\Ngon{4}$ & $\Ngon{4}$ & $\mathbf{15.47\pm0.54}$ & $\mathbf{0.87\pm0.01}$ & $\mathbf{33.96\pm0.18}$ & $\mathbf{2.14\pm0.17}$ \\
    \hline
    P2.1 & $\pi_{n}$ & $\Ngon{4}$ & $\Ngon{5}$ & $15.82\pm0.43$ & $0.87\pm0.01$ & $32.91\pm0.16$ & $1.90\pm0.05$ \\
    P2.2 & $\pi_{\phi}$ & $\Ngon{4}$ ~~ $\Ngon{6}$ & $\Ngon{5}$ & $\mathbf{16.01\pm0.51}$ & $\mathbf{0.89\pm0.02}$ & $\mathbf{32.84\pm0.20}$ & $\mathbf{1.88\pm0.01}$ \\
    \hline
    \end{tabular}
    \caption{\label{tab:widgets} \small Evaluation results of different policies. \textbf{Naming Rule}: IDs started with "C" and "P" means workpieces with 2*N circle pins and 1*1 polygon pin, respectively. \textbf{Experiment Groups}: The policies in C1.x all come from the efficient expert $\pi_e^{\text{eff}}$, while all other experiments are based on the safe expert $\pi_e^{\text{safe}}$. The experiment groups C1.x, C2.x and P1.x compare the expert policy and the cloned policy and the policy learned by RS-GAIL in different situations. The experiment groups C3.x and P2.x compare the nominal policy and the tolerance guided policy in different situations. C4.x compares the policy with and without probabilistic inference. In particular, the policy in C3.1 is identical to the policy in C2.3; the policy in C4.1 is identical to the policy in C3.2. \textbf{Environment Specification}: Our 2*N circle workpieces have 0.3mm pin radius and 0.5mm hole radius, while the interval between 2 rows is 7.62mm, and the intervals between N columns are 2.54mm each. For our 1*1 polygon environments, the pin circumradious and hole circumradious of $\Ngon{4}$ ~~, $\Ngon{5}$ ~~, $\Ngon{6}$ ~~ are (1.5mm, 1.65mm), (1.5mm, 1.65mm), (1.2mm, 1.26mm), respectively.    }
    \label{table:2*nCirclePerformance}
    \vspace{-20pt}
    \end{table}
%===============================================================================
\paragraph{Discussion\label{sec: discussion}}
    %============= Extendability
    The tolerance model can be extended in multiple directions. This paper ignored the rotation outside the xy plane and constructed $\mathcal{T}$ only using the truncated state $[x,y,\theta]$. It is possible to relax the assumption and include more degrees of freedom in the truncated states. Meanwhile, although this paper only considers insertion with spacial tolerances, the proposed method can apply to insertions with tight fit using a tolerance model on force and torque, e.g., $\bar{s}=[x, y,\theta,F_x,F_y,\tau_{\theta}]$. 
    In addition, there are several limitations of the method and will be addressed in future work. First, the reliability of the task encoding is determined by the prediction error of auto-encoder. The accuracy of prediction drops significantly when the tolerance is big in $\theta$ (e.g. over $0.5$ rad). %The limitation is observed during encoding of single pin-hole insertions since tolerance of these tasks is likely to be exceedingly large in $\theta$ direction. 
    We may introduce regularization or normalization of $\theta$ to address the problem. 
    %Second, the probabilistic inference is still computationally expensive, which is not ideal for real-time applications. It now takes at least several seconds to generate the optimal reference. We may develop new efficient optimization techniques to accelerate the computation. 
    Second, to apply the proposed method on real robot hardware, we need to resolve the sim-to-real gap. One method is to use a robust controller \cite{tang2019disturbance} on the robot hardware so that the closed loop dynamics (robust controller plus the hardware) are almost identical to the dynamics used in the simulation. %Another approach is to combine the proposed method with an adaptive controller \cite{johannsmeier2019framework} to directly identify the system dynamics and update the policy in real time.

%===============================================================================

\section{Conclusion}
\label{sec: conclusion}
    This paper introduced a tolerance-guided policy learning method to generate adaptable and transferable policies for delicate industrial insertion tasks. We addressed two practical challenges: generalization of the learned policy to unseen tasks and adaptation of the policy when the workpiece has defects. The proposed method contains three parts. First, we used tolerance embedding to encourage the generalization of the learned policy to different tasks. Second, to deal with efficient training in sparse reward environments, we proposed the reward shaping generative adversarial imitation learning algorithm that leverages the advantages of imitation learning and reinforcement learning. Lastly, we developed probabilistic inference methods to infer optimal insertion points based on failed insertions to make the policy robust to defects of the workpiece. The proposed method has been extensively validated through simulation. It has been demonstrated that 1) the proposed learning algorithm efficiently learned optimal policies under sparse rewards; 2) the tolerance embedding enhanced the transferability of the learned policy; 3) the probabilistic inference improved the success rate for defective workpieces.

%===============================================================================

% The maximum paper length is 8 pages excluding references and acknowledgements, and 10 pages including references and acknowledgements

\clearpage
% The acknowledgments are automatically included only in the final version of the paper.
\acknowledgments{This work is sponsored by Efort Intelligent Equipment Co., Ltd. }%We would like to thank them for their support to our research and their continuous passion for robotics. }

%===============================================================================

\bibliography{example}  % .bib

\end{document}